\documentclass[11pt]{article}
\usepackage{coling2016}
\usepackage{times}
\usepackage[hyphens]{url}
\usepackage{latexsym}
\usepackage{amsmath}
\usepackage{amssymb}
\usepackage[]{algorithm2e}

\makeatletter
\newcommand{\@BIBLABEL}{\@emptybiblabel}
\newcommand{\@emptybiblabel}[1]{}
\makeatother
\usepackage{hyperref}

\usepackage{graphicx}
\usepackage{booktabs}
\usepackage{multirow}
\usepackage[utf8]{inputenc}
\usepackage[T1]{fontenc}

\newcommand{\comment}[1]{}

\title{Exploring Word Embeddings for Unsupervised Textual User-Generated Content Normalization}

\author{Thales Felipe Costa Bertaglia \\
  ICMC-USP / São Carlos -- Brazil \\
  {\tt thales.bertaglia@usp.br} \\\And
  Maria das Graças Volpe Nunes \\
  ICMC-USP / São Carlos -- Brazil \\
  {\tt gracan@icmc.usp.br} \\}

\date{}

\begin{document}
\maketitle
\begin{abstract}
  Text normalization techniques based on rules, lexicons or supervised training requiring large corpora are not scalable nor domain interchangeable, and this makes them unsuitable for normalizing user-generated content (UGC). Current tools available for Brazilian Portuguese make use of such techniques. In this work we propose a technique based on distributed representation of words (or word embeddings). It generates continuous numeric vectors of high-dimensionality to represent words. The vectors explicitly encode many linguistic regularities and patterns, as well as syntactic and semantic word relationships. Words that share semantic similarity are represented by similar vectors. Based on these features, we present a totally unsupervised, expandable and language and domain independent method for learning normalization lexicons from word embeddings. Our approach obtains high correction rate of orthographic errors and internet slang in product reviews, outperforming the current available tools for Brazilian Portuguese. 
\end{abstract}

\section{Introduction}

The huge amount of data currently available on the Web allows computer-based knowledge discovery to thrive. The growth in recent years of user-generated content (UGC) -- especially the one created by ordinary people  -- brings forth a new niche of promising practical applications \cite{Krumm2008}. Textual UGC, such as product reviews, blogs, and social network posts, often serves as a base for natural language processing (NLP) tasks. This type of content may be used for business intelligence, targeted marketing, prediction of political election results, analysis of sociolinguistic phenomena, among many other possibilities. Despite its wide range of application, UGC is hard for NLP to handle.

Most of the natural language processing tools and techniques are developed from and for texts of standard language \cite{Duran2015}. From basic components of a NLP-based system, such as taggers, to complex tools aiming to tackle more significant problems, there is a reliance on well structured textual information in order to achieve a proper behavior. However, user generated content does not necessarily follow the structured form of standard language. This type of text is often full of idiosyncrasies, which represent noise for NLP purposes. Beyond the context of NLP, textual UGC may also represent an obstacle for end-users. Specially on social networks, where a domain-specific and context reliant language is used, users not familiar to its particularities may have difficulties to fully grasp the expressed content. Considering the aforementioned problems, it is relevant to identify noises in UGC for purposes which include a simple identification of the noise, its typification, eventually its substitution for another word or expression. These actions aim to both enhance NLP tools performance and facilitate end-user text comprehension, besides to provide an overview of the linguistics practices on the web. 

The process of identifying noise and suggesting possible substitutions is known as text normalization. Non-standard words (NSWs) are often regarded as noise, but the precise definition of what constitutes them depends on the application domain. Some examples that might be seen as deviations from standard language and that should be normalized include spelling errors, abbreviations, mixed case words, acronyms, internet slang, hashtags, and emoticons. In general, NSWs are words which properties and meaning cannot be derived directly from a lexicon \cite{Sproat2001}. The term "lexicon" in this context does not necessarily mean the list of words that are formally recognized in a language, but rather a set of words that are considered treatable by the specific application. Therefore, it is not possible to clearly state what is noise and what should be normalized.

In contrast to standardized language, UGC is often informal, with less adherence to conventions regarding punctuation, spelling, and style \cite{Wang2013}. Therefore, it is considerably noisy, containing \textit{ad-hoc} abbreviations, phonetic substitutions, customized abbreviations, and slang language \cite{Hassan2013}. Considering the specificity of such content, traditional techniques -- such as lexicon-based substitution -- are not capable to properly handle with many types of noise. Thus, in order to achieve satisfactory results, it is necessary to deeply analyze UGC and develop specific methods aimed at normalizing it.

Conventional string similarity measures (such as edit distance) are not, by themselves, capable of accurately correcting many errors found in UGC. Abbreviations and shorthands found in informal texts, specially on social networks, may contain a large number of edits often resulting in low string similarity -- discouraging the use of traditional techniques.

More recently, the interest in techniques based on distributed representations of words (also called word embeddings) has increased. These representations are able to generate continuous numeric vectors of high-dimensionality to represent words. The vectors explicitly encode many linguistic regularities and patterns, as well as syntactic and semantic word relationships \cite{Mikolov2013}. Words that share semantic similarity are represented by similar vectors. For example, the result of a vector calculation vec(“King”) - vec(“Man”) + vec(“Woman”) is closer to vec(“Queen”) than to any other word vector. Another strong characteristic of distributed representations is their capability of capturing the notion of contextual similarity -- which is essential for textual normalization. The models used for learning these vectors, such as Skip-grams, are totally unsupervised and can be implemented efficiently. In this work, we exploit this set of advantages, combined with lexical similarity measures, in order to capture contextual similarity and learn normalization lexicons based on word embeddings. 

\section{Related work}

Early work handled text normalization as a noisy channel model. This model consists of two components: a source model and a channel model \cite{Shannon:1948}. It assumes that a signal is transferred through a medium and gets corrupted. The source model indicates the canonical form of the signal, and the channel model represents how the signal gets corrupted. \cite{Brill:2000:IEM:1075218.1075255} defined the spelling correction problem as finding $argmax_w P(w | s)$, being $s$ the canonical word, which was sent by the source model, and $w$ the received corrupted word. Applying Bayes' Theorem, the noisy channel model is obtained as $argmax_w P(s | w)  P(w)$. This model presented significant performance improvements compared to previously proposed models, achieving up to 98\% correction accuracy on well-behaved noisy text. However, this approach requires  supervised training data for both canonical and corrupted words. 

Log-linear models also have been applied as unsupervised statistical models for text normalization. \cite{Yang2013} proposed a model in which the relationship between standard and non-standard words may be characterized by a log-linear model with arbitrary features. The weights of these features can then be trained in maximum-likelihood frameworks. The use of this type of model requires a study of the problem to get the most significant features. From the definition of the features, the training process in conducted to optimize the weights. The advantage of these models is the easy incorporation of new features and the optimization is performed according to an objective function. Although not being highly dependent of resources and context-driven, the log-linear approach requires well-defined features -- which are not easily identifiable in UGC. Another disadvantage is the total reliance on statistical observations on the corpus. Hence, the model does not satisfactorily represents the highly semantic specificity of the noise found in UGC, which can occur with low frequency thus not having a significant statistical impact. Considering these issues, this type of model is not enough to deal with generic domain and high context and semantic dependency found is UGC noise. 

More recently, social media text normalization was tackled by using contextual graph random walks. \cite{Hassan2013} proposed a method that uses random walks on a contextual similarity bipartite graph constructed from n-gram sequences on large unlabeled text corpus to build a normalization lexicon. They obtained a precision of 92.43\% and, using the method as a preprocessing step, improved translation quality of social media text by 6\%. \cite{Han2012} also presented an approach for unsupervised construction of normalization lexicons based on context information. Instead of a graph representation, this approach uses string similarity measures between word within a given context. \cite{Wang2013} proposed a supervised learning technique for learning normalization rules from machine translations of a parallel corpus of microblog messages. They built two models that learn generalizations of the normalization process -- one on the phrase level and the other on the character level. The approach was shown able to improve multiple machine translation systems.  

Our technique is most similar to \cite{Kumar2015a}, since we implement an adaptation of the method presented in the mentioned work. The method proposed by \cite{Kumar2015a} aims to learn distributed representations of words to capture the notion of contextual similarity and subsequently learn normalization lexicons from these representations in a completely unsupervised manner. The lexicons are represented as finite-state machines (FSMs) and the process of normalization is performed by transducing the noisy words from the FSMs. Our work makes use of different distributed representation of words, different scoring function for candidate generation and hash structures (dictionaries) instead of FSMs. We also introduce a method for automatically expanding the learned lexicons.     

Regarding Brazilian Portuguese, some studies have been performed considering noises in specific domains, such as reviews of products \cite{Duran2014}, and some tools have been developed specifically for that same domain. The normalizer described in \cite{Duran2015} is, as far as we know, the only tool for text normalization available for Brazilian Portuguese. The proposed lexicon-based normalizer considers that errors found in UGC are divided into six categories: \textbf{Common misspellings:} context-free orthographic errors, often phonetically-motivated. \textbf{Real-word misspellings:} contextual orthographic errors. Words that are contained in the language lexicon, but are wrong considering the context they appear. \textbf{Internet slang}: abbreviations and expressions often used informally by internet users. \textbf{Case use (proper names and acronyms)}: proper names and acronyms wrongly or not at all capitalized. \textbf{Case use (start of sentence)}: sentences starting with a lower case word. \textbf{Glued words}: agglutinated words that should be split. \textbf{Punctuation}: wrong use of sentence delimiters.

Since a large part of misspellings found in UGC is phonetically-motivated, \cite{Duran2015} proposed a phonetic-based speller for correcting such errors. The speller combines edit distance and several specific phonetic rules for Portuguese in order to generate correction candidates. The correction of internet slang and proper name and acronyms capitalization is based on a set of lexicons. Each lexicon contains many pairs of wrong--correct form of words. The correction is performed by looking up the noisy word in the lexicon and substituting it by the correct version. Despite this technique achieving good results in the product review domain, it is not scalable and is too restricted, since there is no form of automatic lexicon-learning. Therefore, it is not suitable for a generic, domain-free normalizer. The results obtained by \cite{Duran2015} will be further discussed, as they are the main source of comparison for our work. 

Another technique specially designed for Brazilian Portuguese is the one proposed by \cite{deMendoncaAlmeida2016}. The work presents two approaches for dealing with spelling correction of UGC. The first approach makes use of three phonetic modules, composed by the Soundex algorithm, a grapheme-to-phoneme converter and a set of language-specific phonetic rules. The second one combines grapheme-to-phoneme conversion and a decision tree classifier. The classifier is trained on a corpus of noisy text and employs 14 features (including string and phonetic similarity measures) to identify and correct different classes of orthographic errors. The approach achieves average correction accuracy of 78\%, however requires training on an annotated corpus and feature extraction -- making it less scalable than an unsupervised technique.   

\section{Distributed Representation of Words}
\label{treinamento}
Distributed representations of words in a vector space, also known as word embeddings, are able to capture lexical, semantic, syntactic, and even contextual similarity between words. This idea has been recently applied to a wide range of NLP tasks with surprising results \cite{Kumar2015a}. The Skip-gram model, introduced by \cite{Mikolov2013}, brought forth an efficient method for learning high-quality vector representations of words from large amounts of unstructured text data. This model, unlike previous ones, does not involve dense matrix multiplications -- making the training optimized and efficient. 

\begin{figure}[!ht]
  	\caption{The Skip-gram model architecture \protect\cite{Mikolov2013}.}
    \label{figSkipGram}
  	\centering
    \includegraphics[width=0.2\textwidth]{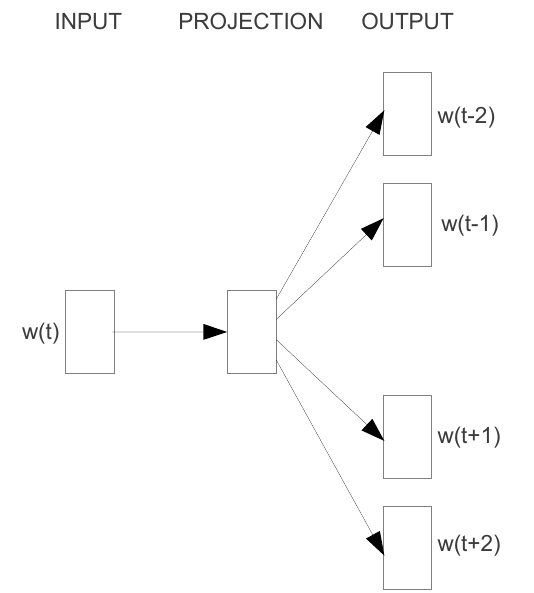}
\end{figure}

The Skip-gram model, graphically represented in Figure \ref{figSkipGram}, tries to maximize classification of a word based on another word in the same sentence. Each current word is used as input to a log-linear classifier and predicts words in a windows before and after the current one. More formally, the goal of the Skip-gram model is to maximize the average log probability given by:

\begin{equation}
\label{logprob}
 \frac{1}{T} \sum\limits_{t=1}^T \sum\limits_{-c\leq j\leq c, j\neq0} \log p(w_{t+j} | w_t) 
\end{equation}
where $c$ is the size of the training context, centered on word $w_t$ \cite{MikolovEff}. In practice, word embeddings can be trained using many frameworks that already implement the training process. We used the Gensim \footnote{\url{https://radimrehurek.com/gensim/index.html}} framework for Python. The process and parameters we used to train the embeddings will be explained in the next section.

For training distributed representations of words, two sources of data were used in our work. One is a large corpus of product reviews containing both noisy and correct texts. A complete description of the corpus can be found in \cite{HARTMANN14.413}. The other one is a corpus of Twitter posts (tweets) written in Portuguese. The tweets were extracted using a crawler developed by the authors and the data cannot be distributed due to license restrictions. Most of that content is noisy, considering that the tweets are limited by a maximum of 140 characters and that Twitter is, in general, an informal medium of content generation. 

Both corpora had to be preprocessed in order to improve the text quality. First, the Twitter corpus was tokenized using Twokenizer \footnote{\url{http://www.cs.cmu.edu/\~ark/TweetNLP/\#pos}}, a specific tokenizer for tweets. User names, hashtags and hyperlinks were removed. The product review corpus was tokenized using a simple word tokenizer and predefined tags contained in the reviews (such as "What I liked about this product:") were removed. Next, the sentences were segmented  by using the tool provided by \cite{Duran2015}. After segmentation, we obtained 6.8 million sentences from the Twitter corpus and 20 million from the product reviews corpus. 

\section{Similarity Measures}

In order to find candidates for normalizing noisy words, some similarity measures were employed. The cosine distance between two D-dimensional vectors $u$ and $v$ can be used to determine how similar are two word embeddings. It is defined as:

\begin{equation}
\text{cosine  similarity} = \frac{ \sum\limits_{i=1}^D u_i \times v_i }{\sqrt[]{ \sum\limits_{i=1}^D (u_i)^2 \times \sum\limits_{i=1}^D (v_i)^2 }}
\end{equation}

To illustrate the representation power of word embeddings, the following image shows the 25 nearest neighbors (closest cosine similarity) of word '\textit{você}' (you) obtained from our word embedding model: 

\begin{figure}[!ht]
  	\caption{25 nearest neighbors for word '\textit{você}' (you) found in our word embedding model.}
    \label{figVoce}
  	\centering
    \includegraphics[width=0.6\textwidth]{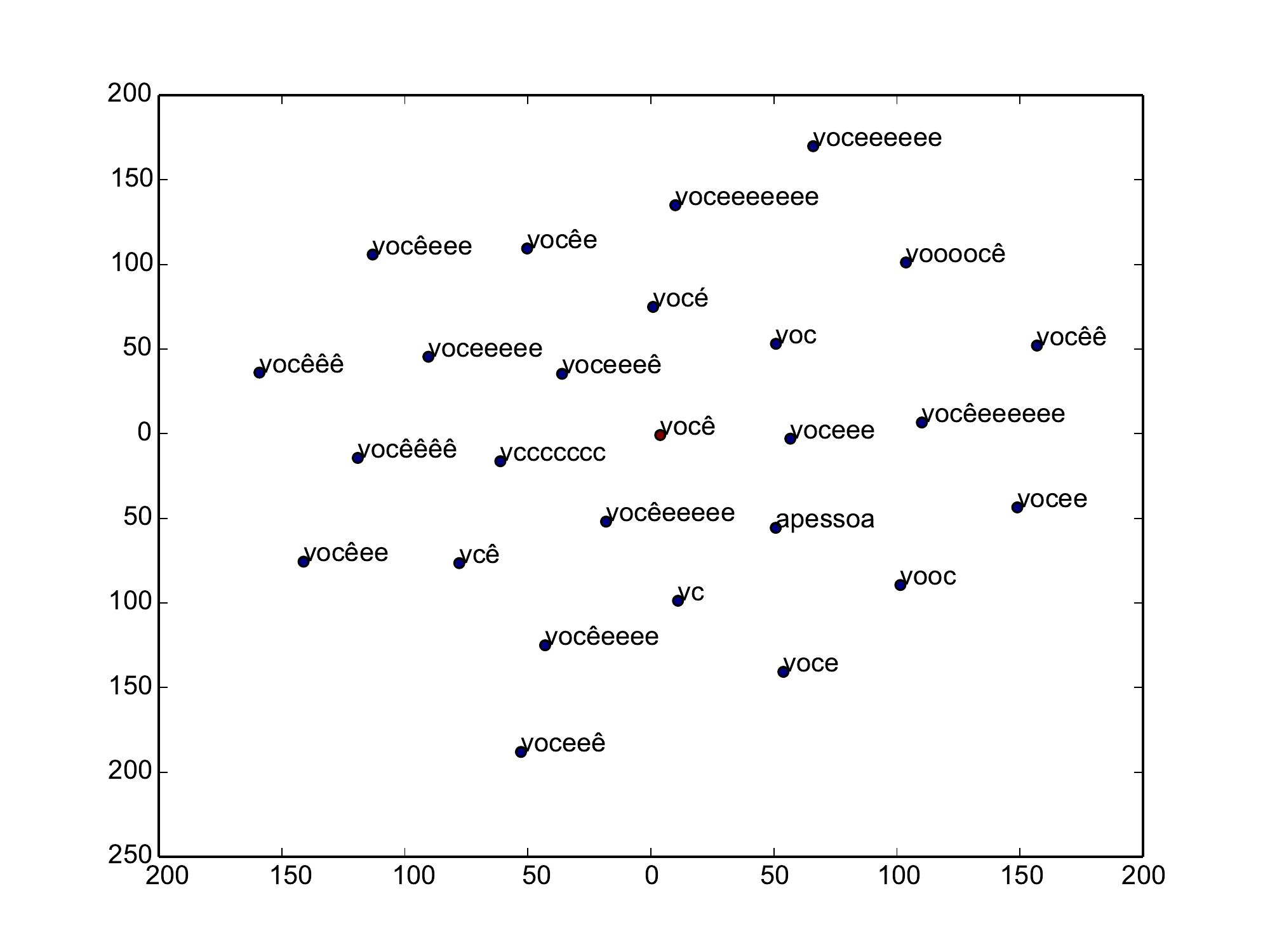}
\end{figure}

In Figure \ref{figVoce}, the canonical form of the word (bold) is located at the center. Around it, there are many noisy variations found in the embedding model. In order to find the canonical form parting from a noisy word, it is also necessary to consider the lexical similarity. We compute its value using an adaptation of the definition presented in \cite{Hassan2013}, as:

\begin{equation}
\text{lexical similarity}(w_1, w_2) =
\begin{cases}
\frac{LCSR(w_1,w_2)}{MED(w_1,w_2)}, & \text{if} MED(w_1,w_2) > 0 \\
LCSR(w_1,w_2), & \text{otherwise}
\end{cases}
\end{equation}
where $w_1$ and $w_2$ are two given words and $MED(w_1,w_2)$ is the modified edit distance between them, calculated as $MED(w_1,w_2) = ED(w_1,w_2) - DS(w_1, w_2)$. $DS(w_1, w_2)$ is the \textit{diacritical symmetry} between $w_1$ and $w_2$ -- \textit{i.e}, the number of characters from a word that are aligned with versions of itself with diacritical marks in the other word. For instance, the diacritical symmetry between \textit{maçã} (apple) and \textit{maca} (stretcher) is 2. This measure is employed due to the fact that many misspellings in Portuguese happen because of a single diacritic, thus this modification improves the correction of such errors.  $LCSR$ refers to Longest Common Subsequence Ratio and is calculated, in our modified version, as:

\begin{equation}
LCSR(w_1, w_2) = \frac{LCS(w_1,w_2) + DS(w_1, w_2)}{\text{max length}(w_1,w_2)}
\end{equation}
where $LCS$ refers to Longest Common Subsequence, being usually obtained through a dynamic programming approach. The lexical similarity measure indicates how similar two words are, ranging from 0 (no similarity) to 1 (identical).

\section{Learning Normalization Lexicons}

Having obtained the set of word embeddings, the method requires a list of canonical words as input. This list may be a lexicon of words recognized by a language, but may also be a list of words considered treatable by the application. We used the UNITEX-PB \footnote{\url{\detokenize{http://www.nilc.icmc.usp.br/nilc/projects/unitex-pb/web/dicionarios.html}}} lexicon for Brazilian Portuguese, removing words with frequency below 100 on Corpus Brasileiro \footnote{\url{http://corpusbrasileiro.pucsp.br/cb/Acesso.html}} frequency list. This pruning step removes words that probably do not appear in UGC, reducing the lexicon size and improving performance. The next step obtains the K-nearest neighbors, excluding words contained in the lexicon, in the vector space for each canonical word contained in the lexicon. We used K = 25, following \cite{Kumar2015a} experiments. This step creates a list of possible corruptions (noisy versions) for every canonical word $w_c$. Considering the vector space $V$, the lexicon $L$ and $w_c$ every word $\in L$, the list of noisy versions is obtained as:

\begin{equation}
\text{noisy versions}[w_c] = \forall_{w \in V,V\notin L}\max_{k}\text{cosine similarity}(w_c, w) 
\end{equation}

After obtaining the list of noisy versions, it is necessary to invert it, because we must find the canonical version given a noisy word -- therefore, the noisy version must index a list of correction candidates to it. In order to do so, we loop through every noisy versions list. We create a new structure to store each noisy word $w_n$ as index to a list of canonical words that contain $w_n$ in its noisy versions list. We then store a tuple $(w_c,score(w_n,w_c))$ as a candidate to correct $w_n$. The score is calculated as:

\begin{equation}
score(w_n, w_c) = n\times \text{lexical similarity}(w_n,w_c) + (1-n)\times \text{cosine similarity}(w_n,w_c) 
\end{equation}

Different from \cite{Kumar2015a}, our approach considers the cosine similarity at the score function, because some corrections have low lexical similarity but appear in the same context, specially for abbreviations -- for instance, \textit{d+} as abbreviation for \textit{demais} (too much). The term $n$ is a value between 0 and 1 that indicates the weight (or importance) given to the lexical similarity. In our experiments, $n$ was set as $0.8$. The learned lexicon is stored as a hash structure, indexed by $w_n$, where each entry contains the list of candidate corrections for the respective noisy word. To determine the best candidate to correct $w_n$, we simply compute $\max(score(w_n,w_c))$. Algorithm \ref{alg} summarizes the lexicon learning process.

\begin{algorithm}[H]
 \textbf{Input:} {list of canonical words $L$, word embedding model $V$, number of $K$ nearest neighbors}\\
 \textbf{Output:} {learned lexicon $T$ where $T[w_n]$ is the list of candidates for correcting noisy word $w_n$ }\\
 \For{each $w \in L$}
 {
  	\For{each $v \in V$}
    {
  		\If{$v\notin L$}{compute cosine similarity$(w,v)$}
  		store top $K$ neighbors in $C(w)$
  	}
 }
 \For{each $w\in C$}
 {
 	\For{each $w' \in C[w]$}
    {
    	$T[w'] \Leftarrow (w, score(w',w))$
    }
 }
 \caption{Unsupervised Lexicon Learning}
 \label{alg}
\end{algorithm}

\subsection{Expanding the Learned Lexicon}

Given that only the $K$-nearest neighbors of a canonical word are added to the lexicon, some very infrequent noisy words may not be a top neighbor of any canonical word. In this case, the approach presented above cannot find any correction candidates. In order to solve this issue, we added an expansion step to the method. If a given noisy $w_n$ is not in the learned lexicon, that is, $w_n \notin T$, then we find the canonical word from $L$ which is most similar to $w_n$, and add it as a correction candidate. Thus, the expansion step can be defined as:

\begin{equation}
\text{If } w_n \notin T: T[w_n] \Leftarrow \max_{w_c \in L}\text{lexical similarity}(w_n, w_c) 
\end{equation}

Despite its simplicity, the expansion step improved the overall correction recall.

\subsection{Representing Context}

Thus far, our framework has scored candidates deterministically, based solely on lexical information. In this case, correction candidates will always be the same for a given noisy word, independent from the context it appears. This idea might be enough for correcting internet slang, since equal abbreviations are seldom used for different words. However, ignoring contextual information makes it impossible to handle contextual orthographic errors (also called real-word errors or RWEs). This type of error occurs when a given word is in the recognized lexicon (so it is not \textit{per se} a noisy word) but is incorrect due to the context it appears. In Portuguese, cases of real-word errors caused by missing a single diacritic are frequent and can only be corrected by considering the context. In order to do so, we used a language model (LM) (with trigrams) trained on a Wikipedia sample consisting of 3841834 sentences. We trained our LM using the KenLM framework, which employs modified Kneser-Ney smoothing for estimation \cite{kenlm}. The estimated trigram probabilities were normalized to fit the range of the other measures and then added to the scoring function. The correction of RWEs relies only on similarity measures and on the language model, because it maps a canonical (but inadequate to the context) word to another canonical word -- therefore the learned lexicon based on word embeddings does not include it. 

\section{Evaluation}

In order to evaluate the proposed method and compare it to already existing tools for Brazilian Portuguese, two annotated samples of product reviews written by users were used. Each sample contains 60 reviews, with every error manually annotated by a specialist. The annotation considers the six categories of noise proposed by \cite{Duran2015}, but our technique can only be applied to the correction of orthographic errors and internet slang.  

First, we conducted experiments in order to determine the best word embedding model. We trained both Skip-gram and Continuous bag-of-words (Cbow) models, implemented in Gensim, to learn the embeddings. We used a context window of size 5, \textit{i.e} 2 words before and 2 after the center, and used hierarchical sampling for reducing the vocabulary size during training, considering only words with a minimum count of 10 occurrences. We generated embeddings with 100, 300 and 500 dimensions. For each dimension size, we trained the embeddings on two different sets of data: the first one, referred as Noisy, is exactly as described in Section \ref{treinamento}, and the second one, referred as Hybrid, includes an additional dataset containing 38 million sentences from Wikipedia. Table \ref{tableEmbs} shows the recall measure obtained for the correction of orthographic errors (O) (without the LM probabilities) and internet slang (I) (with the best setup) using each embedding model trained. 

\begin{table}[!h]
\centering
\caption{Results for error correction obtained from each word embedding model}
\label{tableEmbs}
\begin{tabular}{@{}lllllll@{}}
\toprule
\multicolumn{1}{c}{\textbf{}}       & \multicolumn{6}{c}{\textbf{Dimensions}}                                                                                                             \\ \cline{2-7}
\multicolumn{1}{c}{\textbf{Model}} & \multicolumn{2}{c}{100}                        & \multicolumn{2}{c}{300}                        & \multicolumn{2}{c}{500}                        \\ \cline{2-7}
\multicolumn{1}{c}{\textbf{}}       & \multicolumn{1}{c}{O} & \multicolumn{1}{c}{I} & \multicolumn{1}{c}{O} & \multicolumn{1}{c}{I} & \multicolumn{1}{c}{O} & \multicolumn{1}{c}{I} \\ \midrule
Cbow Hybrid                           & 67.2\%                      & 64.5\%                      & 72.3\%                      & 64.5\%                      & 72.1\%                      & 64.5\%                      \\ \midrule
Cbow Noisy                           & 70.7\%                     & 54.8\%                      & 74.7\%                      & 64.5\%                      & 76.0\%                      & 64.5\%                      \\ \midrule
Skip-gram Hybrid                      & 63.8\%                      & 77.4\%                      & 74.2\%                      & 71.0\%                      & 76.0\%                      & 71.0\%                      \\ \midrule
Skip-gram Noisy                      & 77.3\%                      & 64.5\%                      & 78.6\%                      & 67.7\%                      & \textbf{78.6\%}                      & \textbf{77.4\%}                      \\ \bottomrule
\end{tabular}
\end{table}
The Skip-gram model with 500 dimensions showed the best results. Therefore, further experiments were conducted using these embeddings. In order to better evaluate the framework, we performed experiments using three different learned lexicons models. The only difference between them is the word embedding model employed. The first one, referred as \textit{Noisy}, uses embeddings trained on data preprocessed exactly as described in Section \ref{treinamento}. The second model, referred as \textit{Clean}, uses embedding trained on data with an additional preprocessing step: every symbol that is not either a letter or a number is removed. Both are Skip-gram models with 500 dimensions. The third model is an  \textit{Ensemble} containing the previous ones. Its output is defined as $max(output(Clean),output(Noisy))$. We have found empirically that the \textit{Noisy} model is better for internet slang correction, the \textit{Clean} model is better for orthographic errors and the \textit{Ensemble} joins the best of both. In Tables \ref{resOrt} and \ref{resInt} we compare the three lexicon models with UGCNormal, the tool proposed by \cite{Duran2015}. The tables follow the format $X/Y=Z$, being X the total of corrections performed by each model, Y the total of annotated errors in the sample and Z the obtained recall measure.
\begin{table}[!h]
\centering
\caption{Results for orthographic errors correction}
\label{resOrt}
\begin{tabular}{lc}
\toprule
\bf Model      					& \bf Corrections 					\\ \midrule
Noisy                            & 129/164 = 78.6\%                 \\ \midrule
Clean                            & 135/164 = 82.3\%                 \\ \midrule
Ensemble                         & 137/164 = 83.5\%                 \\ \midrule
Ensemble+Expansion Step 		 & 149/164 = 90.9\%        			\\ \midrule
Ensemble+Expansion Step+LM 		 & \textbf{151/164 = 92.1\%}        \\ \midrule
Expansion Step+LM (RWEs) 		 & \textbf{84/115 = 73.0\%}        \\ \midrule
UGCNormal                        & 137/164 = 83.5\%                 \\ \midrule
UGCNormal (RWEs)                 & 39/115 = 33.9\%                 \\ \bottomrule
\end{tabular}
\end{table}

\begin{table}[!h]
\centering
\caption{Results for internet slang correction}
\label{resInt}
\begin{tabular}{lc}
\toprule
\bf Model      					 	& \bf Corrections 				   \\ \midrule
Noisy                            	& 20/31 = 64.5\%                   \\ \midrule
Clean                            	& 17/31 = 54.8\%                   \\ \midrule
Ensemble                         	& 22/31 = 71.0\%                   \\ \midrule
Ensemble+Expansion Step 		 	& 23/31 = 74.2\%         		   \\ \midrule
Ensemble+Expansion Step+LM 			& \textbf{24/31 = 77.4}\%          \\ \midrule
UGCNormal                       	& 19/31 = 61.3\%                   \\ \bottomrule
\end{tabular}
\end{table}
The results show that the ensemble model with expansion step and context representation (LM) outperforms every other model, including UGCNormal. For RWEs, the combination of expansion step and language model outperforms UGCNormal by a large margin.  

\section{Conclusion}
We presented an unsupervised method for learning normalization lexicons based on similarity measures and distributed representation of words. Our approach does not require sets of rules or domain-specific lexicons, but only a lexicon containing a list of canonical words. Therefore, it can be easily expanded and adapted according to the needs of each specific application. We trained distributed representations using the Skip-gram model on a large corpus of both Twitter posts and product reviews. The results indicate that the method for both internet slang and orthographic error correction surpasses the results obtained from already existing tools for Brazilian Portuguese. The expansion step, despite being simple, improved the correction significantly.  As future work, different methods of context representation must be explored. Methods for dealing with multiword expressions, such as acronyms for internet slang, must be added to our framework. Despite being rare in Brazilian Portuguese, such language constructs are commonly find on internet slang, mainly derived from English.  Correcting more categories of noise, either by expanding this technique or investigating new approaches, is the following step for this project.

\bibliographystyle{acl}
\bibliography{coling2016}

\end{document}